\documentclass[runningheads]{llncs}
\usepackage[T1]{fontenc}
\usepackage[misc]{ifsym}
\usepackage{booktabs}
\usepackage{graphicx}

\usepackage{amssymb,amsmath,array}
\usepackage{tikz}
\usetikzlibrary{arrows,shapes}
\usetikzlibrary{patterns}
\usetikzlibrary{decorations.pathreplacing, patterns}
\usepackage{tabularx}
\usepackage{multicol}
\usepackage{multirow}

\usepackage{algorithm}
\usepackage{algorithmicx}
\usepackage{algpseudocode}
\usepackage{hyperref}
\usepackage{listings}
\newfloat{algorithm}{t}{lop}
\algnewcommand\algorithmicsymbols{\textbf{Symbols:}}
\algnewcommand\ASymbols{\item[\algorithmicsymbols]}
\algnewcommand\algorithmicinput{\textbf{Input:}}
\algnewcommand\algorithmicoutput{\textbf{Output:}}
\algnewcommand\Input{\item[\algorithmicinput]}
\algnewcommand\Output{\item[\algorithmicoutput]}
\algnewcommand\algorithmicforeach{\textbf{for each}}
\algdef{S}[FOR]{ForEach}[1]{\algorithmicforeach\ #1\ \algorithmicdo}

\begin{document}

\title{Transforming Datasets to Requested Complexity with Projection-based Many-Objective Genetic~Algorithm}
\titlerunning{Transforming Datasets to Requested Complexity}

\author{Joanna Komorniczak}

\institute{Wrocław University of Science and Technology \\
Wyb. Wyspiańskiego 27, 50-370 Wrocław, Poland}

\authorrunning{J. Komorniczak}

\maketitle              

\begin{abstract}
The research community continues to seek increasingly more advanced synthetic data generators to reliably evaluate the strengths and limitations of \textit{machine learning} methods. This work aims to increase the availability of datasets encompassing a diverse range of problem \textit{complexities} by proposing a \textit{genetic algorithm} that optimizes a set of \textit{problem complexity measures} for \textit{classification} and \textit{regression} tasks towards specific targets. For classification, a set of 10 complexity measures was used, while for regression tasks, 4 measures demonstrating promising optimization capabilities were selected. Experiments confirmed that the proposed genetic algorithm can generate datasets with varying levels of difficulty by transforming synthetically created datasets to achieve target \textit{complexity} values through linear feature projections. Evaluations involving state-of-the-art classifiers and regressors revealed a correlation between the complexity of the generated data and the recognition quality.

\end{abstract}

\section{Introduction}

In today’s era of widespread applications of \textit{machine learning}, considering both the risks and hopes associated with its rapid development and growing popularity \cite{kaur2022trustworthy}, one must pay particular attention to the reliable and accurate evaluation of proposed methods. In many cases, using real-world datasets for method evaluation appears to confirm the effectiveness of a specific algorithm and demonstrates its usability \cite{stapor2021design}. However, the limited availability of diverse real-world datasets is a commonly recognized issue in machine learning research \cite{lu2023machine}. 

The use of synthetic data offers many benefits, including reduced storage requirements, as large amounts of data can be generated using a synthetic strategy and a specific random state. What is more, data with particular characteristics can be requested -- i.e., with large dimensionality or specific relationships between features. When evaluating novel methodologies, demonstrating an increase in recognition performance on challenging datasets is usually considered desirable. Specifying hyperparameters of a generation method enables one to generate data that poses various challenges. Such challenges may be related to data dimensionality, feature informativeness, noise level, or class imbalance \cite{guyon2007competitive}. 

\subsection{Synthetic datasets in \textit{machine learning} research}

One of the most frequently used classification data synthesis methodologies is~\textit{Madelon} \cite{guyon2007competitive}, implemented in a \verb|make_classification| function of \textit{scikit-learn} library. The generator samples data distributions lying on the vertices of a multidimensional hypercube. A particular benefit of this approach is the possibility of specifying the number of \textit{informative}, \textit{redundant}, and \textit{repeated} features, as well as the length of the hypercube edge related to class separation.

Some other generation strategies focus on specific difficulties related to class overlaps across features. An example may be the two moons dataset \cite{van2020uncertainty} or \textit{spiral} and \textit{circles} datasets, where classes are not linearly separable~\cite{ben2021sparsity}. In general, synthetic data generation methods aim to produce datasets that test the capabilities of recognition methods. For example, Mitra et al. \cite{mitra2024variable} proposed 10 strategies for generating classification data to evaluate methods' feature-extraction abilities. Other research used data generated based on logic gates or LED displays \cite{kamalov2023synthetic}. Another reason for using synthetic data is to preserve confidential and sensitive information. In that case, some data generation methods aim to approximate the real-world sample distribution by employing \textsc{smote} or deep learning methods like Generative Adversarial Networks or Variational Autoencoders \cite{endres2022synthetic}. Such a generator aims to reliably resemble the actual real-world data \cite{drechsler2011empirical}. 


In a regression task, synthetic data can be generated by randomly sampling features and then calculating the target variable based on their values. \textit{Friedman} datasets \cite{breiman1996bagging} describe three types of non-linear relations between features and a target using trigonometric functions and feature multiplications. A particularly interesting synthetic dataset for regression and dimensionality reduction is the \textit{swissroll} dataset \cite{marsland2011machine}, which describes points lying on a 3D non-linear structure. A commonly used generation strategy, implemented in the \verb|make_regression| function of the \textit{scikit-learn} library, uses linear regressors on a random feature matrix and Gaussian noise with a specified level to produce the target variable.

\subsection{Complexity measures}

Complexity measures aim to estimate the difficulty of a given recognition problem. Those measures have been proposed and systematized for classification \cite{lorena2019complex} and regression tasks \cite{lorena2018data}. They usually serve as tools for dataset description in experimental research \cite{11073800}, since various measures and their combinations can distinguish simple problems (e.g., linearly separable) from more complex ones \cite{maciel2016measuring,scholz2021comparison}. Some research uses complexity measures in the preprocessing stage. Such applications include selecting data resampling ratios for imbalanced datasets \cite{lancho2025selecting} or noise filtering \cite{saez2013predicting}.

Given the benefits of synthetic data, the complexity metafeatures can serve as a valuable tool for data generation. One of the first approaches of that kind used a genetic algorithm that searched for the combination of labels that optimized the \textit{fraction of borderline points} (N1) complexity measure \cite{macia2008genetic}. The 2018 research used a \textit{greedy search} algorithm to optimize the F1, N1, and N3 measures individually \cite{de2018using}, also by assigning existing samples to different classes. Later approaches focused on multi- and many-criteria optimization with sophisticated optimization algorithms like NSGA-III \cite{deb2013evolutionary}, extending the set of optimized measures to three (N1, N2, and F1) \cite{macia2009beyond} or four (C1, L2, N1, F2) \cite{fracca2020many}. Some approaches adjusted the labels of real-world data distributions to achieve the specific problem complexity that resembles that of real-world data \cite{9981848}. In all of the mentioned research, the individuals in an evolutionary algorithm were the vectors describing the class labels of a specific sample distribution.

A different approach was proposed in the \textit{Sy:Boid} method \cite{houston2023genetically}, where the authors adjusted the \textit{boids} simulation, originally intended to model the movement of birds for computer graphics, to produce synthetic datasets with varying complexities. This approach used a genetic algorithm to search over the hyperparameters of the generation method, aiming to achieve a specified level of complexity. This approach optimized the F1 and N1 measures.

\subsection{Contribution}

This work introduces a different approach to altering the \textit{complexity} of a dataset, which modifies the data distribution via \textit{feature projections} using a transformation matrix whose coefficients are optimized via a genetic algorithm. In the presented research, the source distribution is synthetically generated. By default, the dataset's dimensionality $d$ remains unchanged; hence, each individual in the population is a $d \times d$ transformation matrix. 

\begin{figure}[!b]
    \centering
    \includegraphics[width=\linewidth]{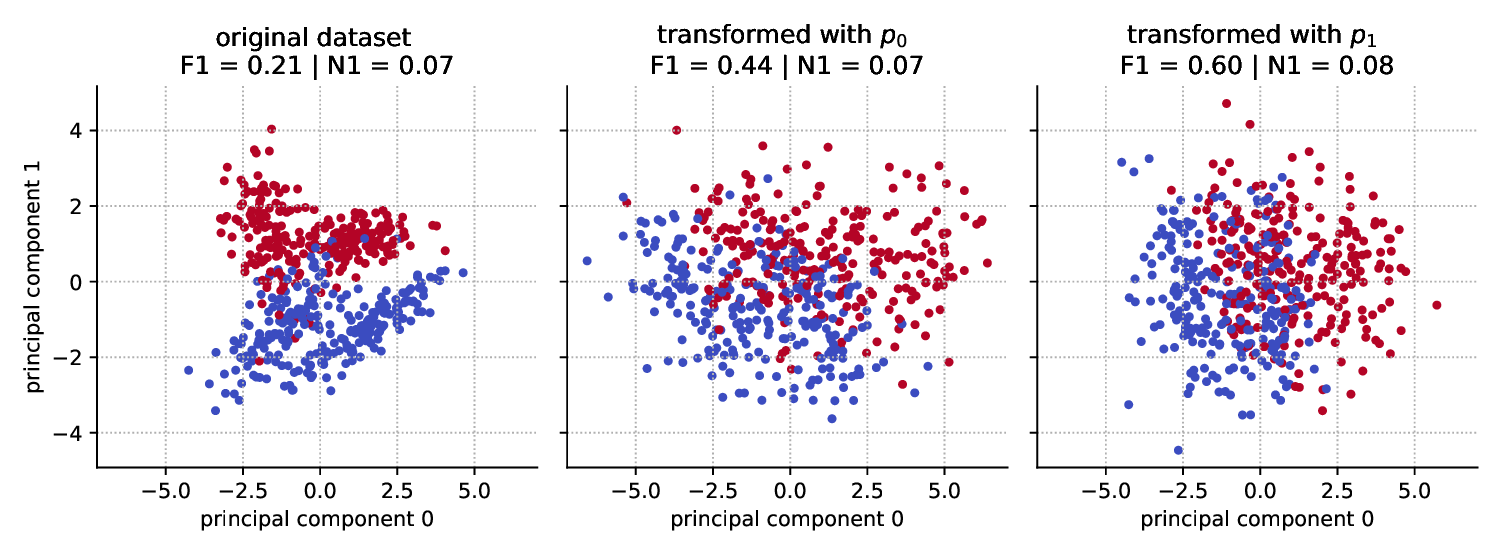}
    \caption{Principal components of the dataset produced using random feature projections of a 10-dimensional synthetic classification problem. The original data is shown in the first plot, and the results of feature transformations in the following ones.}
    \label{fig:exaple_projection}
\end{figure}

An example of feature projections using two different random transformation matrices $p_0$ and $p_1$ for a synthetic dataset with a dimensionality $d=10$ is presented in Figure \ref{fig:exaple_projection}. The first subplot presents the \textsc{pca} components of the original, unmodified dataset. The other two subplots show the principal components of a dataset after applying random projection. The complexity measures calculated for the datasets, displayed in the figure titles, indicate that the complexity of the dataset increased after applying the projection.

In summary, this work presents a \textit{genetic algorithm} that transforms datasets to a specified level of \textit{complexity}. The proposed solution enables the use of a synthetically generated dataset for \textit{classification} or \textit{regression} tasks and allows for adjusting the \textit{problem complexity} level toward a desired target specified by hyperparameters. Compared to existing strategies, the search space proposed in this approach is \textit{continuous} and \textit{unbounded}. 

The main contributions of the presented research are as follows:
\begin{itemize}
    \item Proposition and presentation of an \textit{Evolutionary Projection-based Complexity Optimization} (\textsc{epco}) optimizing the problem complexity measures to~a~desired target for \textit{classification} and \textit{regression} tasks.
    \item Selection of complexity measures for the considered tasks that show promising optimization possibilities via feature projections.
    \item Transformation of synthetically generated datasets to five different complexity levels using the proposed approach.
    \item Experiments investigating how altering the problem complexities affects the recognition quality for classification and regression tasks using state-of-the-art methods.
\end{itemize}

\section{Method}

This article presents the \textit{Evolutionary Projection-based Complexity Optimization} (\textsc{epco}) -- a genetic algorithm that optimizes a set of predefined criteria, $\mathcal{C} = \{C_0, C_1, \dots, C_N\}$ describing the complexity of a classification or regression problem, towards a target level specified with a hyperparameter. For each criterion $C_i$, a fitness function is defined as $F_i = |T_i - C_i|$, where $C_i$ is the value of the measure for a given individual, and $T_i$ is the desired target value of this measure. The goal is to minimize $F_i$.

\subsection{Algorithm description}

The operation of the proposed approach is presented in Algorithm \ref{alg:pseudo}. The method receives a {source $d$-dimensional dataset $\{X, y\}$ as a parameter, as well as the set of complexity measures to optimize $\mathcal{C}$ with their target values $\mathcal{T}$. The original dataset $X$ contains $n$ samples $X = \{x_0, x_1, \dots, x_{n-1}\}$, and each sample $x_i$ is a $d$-dimensional vector $x = [x^0, x^1, \dots, x^{d-1}]$. The $n$-dimensional label vector $y$ describes the continuous or discrete target for each sample $x_i$. The remaining hyperparameters describe the evolutionary algorithm: population size $m$, number of iterations $I$, crossover and mutation ratios $r_c$ and $r_m$, as well as the decay factor $d_f$, which reduces the number of crossover and mutation operations at a later stage of evolution.

\begin{algorithm}[!t]
\scriptsize
\caption{\emph{Evolutionary Projection-based Complexity Optimization}}
\label{alg:pseudo}
\begin{algorithmic}[1]
\Input
\Statex $\{X, y\}$ -- source dataset, $\mathcal{C}$ -- set of complexity measures to optimize, $\mathcal{T}$ -- list of target values of measures, $m$ -- population size, $I$ -- number of iterations, $r_c$ -- crossover ratio, $r_m$ -- mutation ratio, $d_f$ -- decay factor
\Output
\Statex $\{X_p, y\}$ -- output dataset
    \Statex \Comment{\textit{initialize population and measure fitness}}
    \State $\mathcal{P} \gets m$ matrices of size $d$ x $d$ with random values drawn from $\mathcal{N}(0,3)$    
    \State $\mathcal{F} \gets \emptyset$
    \ForEach{$p_i$ \textbf{in} $\mathcal{P}$}
        \State $\mathcal{F}_{i} \gets$ calculate fitness of $X \otimes p_i$ 
    \State $\mathcal{F} \gets \mathcal{F} \cup \mathcal{F}_i$
    \EndFor
    \Comment{\textit{\textbf{iterate through evolutionary process}}}
    \ForEach{$i$ \textbf{in} $\{ 0, 1, 2, \ldots I\}$}
    \Comment{\textit{order population according to fitness}}
    \State $\mathcal{P} \gets$ order samples according to $\mathcal{F}$
    \State $\mathcal{F} \gets$ order fitness according to $\mathcal{F}$
    \Comment{\textit{\textbf{crossover}}}
    \State $n_c \gets max(1, m \cdot c_r)$
    \ForEach{$i_c$ \textbf{in} $\{ 0, 1, 2, \ldots n_c\}$}
        \State $s_1 \gets \mathcal{P}[i_c]$
        \State $s_2 \gets $ random individual from $\mathcal{P}$
        \State $p' \gets crossover(s_1, s_2)$
        \Comment{\textit{substitute individuals with worst fitness}}
        \State $\mathcal{P}[-i_c] \gets p'$
        \State $\mathcal{F}_{-i_c} \gets$ calculate fitness of $X \otimes p'$
    \EndFor
    \Comment{\textit{\textbf{mutation}}}
    \State $n_m \gets m \cdot c_m$
    \ForEach{$i_m$ \textbf{in} $\{ 0, 1, 2, \ldots n_m\}$}
        \State $s \gets $ random individual position
        \State $p' \gets mutate(p_s)$
        \State $\mathcal{F}_{s} \gets$ calculate fitness of $X \otimes p'$
    \EndFor
    \Comment{\textit{reduce crossover and mutation ratios}}
    \State $r_c \gets r_c \cdot (1-d_f)$ 
    \State $r_m \gets r_m \cdot (1-d_f)$ 
    \EndFor
    \Comment{\textit{return dataset transformed with leading projection}}
    \State $X_p \gets X \otimes p_{|\mathcal{C}|}$
    \State \textbf{return} $\{X_p, y\}$
\end{algorithmic}
\end{algorithm}

\paragraph{Initialization}
The procedure starts with initializing the population. Each individual is a transformation matrix of size $d \times d$, where $d$ describes the data dimensionality. The coefficients in transformation matrices are drawn from the normal distribution $\mathcal{N}(\mu,\sigma)$ with a mean value of $\mu = 0$ and a standard deviation of $\sigma=3$. In the next stage of processing, the individuals are scored based on the absolute differences between the criteria $\mathcal{C}$ and their target values $\mathcal{T}$. First, each projection in the population $p_i$ is used to produce a new feature set $X' = X \otimes p_i$, where $\otimes$ describes matrix multiplication and $X$ the original features. Later, the fitness in a given criterion is calculated as $|T_j - C_j(X', y)|$ for a measure $C_j$ and its target $T_j$. This is described in the lines \verb|2:6| of the pseudocode.

\paragraph{Natural selection}
The optimization runs for a specified number of iterations. The size of a population remains constant, with the order of individuals in a population following their \textit{fitness} scores. The individuals are sorted according to fitness in each criterion $C_i$ and the sum of their fitness scores, $\Sigma_C$. Then, the population is reordered in such a way that the first $|\mathcal{C}|$ individuals have the highest scores in consecutive criteria, and the individual in position $|\mathcal{C}|+1$ has the highest score in the sum of the criteria. Ordering individuals by fitness allows a straightforward approach to \textit{selection}, replacing poorly adapted individuals at the tail of the population, and serves as a solution for selecting the best individuals for crossover. The illustrative example of such an order is presented in Figure \ref{fig:order} for three complexity measures. 

\begin{figure}[t]
		\centering
    	\resizebox{0.8\textwidth}{!}{ \begin{tikzpicture}
  \node at (5.5,2) {\textbf{position in population}};
  \node at (5.5,-2) {\textbf{primary criterion}};

  \draw[decorate, decoration={brace, amplitude=6pt}, thick]
    (-0.5,1.2) -- (11.5,1.2);

  \draw[decorate, decoration={brace, amplitude=6pt, mirror}, thick]
    (-0.5,-1.4) -- (11.5,-1.4);

  \node[draw, pattern=vertical lines, pattern color=black, minimum size=1cm] at (0,0) {};
  \node at (0,0.75) {0};
  \node at (0,-0.75) {$C_0$};

  \node[draw, pattern=horizontal lines, pattern color=black, minimum size=1cm] at (1,0) {};
  \node at (1,0.75) {1};
  \node at (1,-0.75) {$C_1$};

  \node[draw, pattern=north east lines, pattern color=black, minimum size=1cm] at (2,0) {};
  \node at (2,0.75) {2};
  \node at (2,-0.75) {$C_2$};

  \node[draw, pattern=north west lines, pattern color=black, minimum size=1cm] at (3,0) {};
  \node at (3,0.75) {3};
  \node at (3,-0.75) {$\Sigma_C$};

  \node[draw, pattern=vertical lines, pattern color=black, minimum size=1cm] at (4,0) {};
  \node at (4,0.75) {4};
  \node at (4,-0.75) {$C_0$};

  \node[draw, pattern=horizontal lines, pattern color=black, minimum size=1cm] at (5,0) {};
  \node at (5,0.75) {5};
  \node at (5,-0.75) {$C_1$};

  \node[draw, pattern=north east lines, pattern color=black, minimum size=1cm] at (6,0) {};
  \node at (6,0.75) {6};
  \node at (6,-0.75) {$C_2$};

  \node[draw, pattern=north west lines, pattern color=black, minimum size=1cm] at (7,0) {};
  \node at (7,0.75) {7};
  \node at (7,-0.75) {$\Sigma_C$};

  \node[draw, pattern=vertical lines, pattern color=black, minimum size=1cm] at (8,0) {};
  \node at (8,0.75) {8};
  \node at (8,-0.75) {$C_0$};

  \node[draw, pattern=horizontal lines, pattern color=black, minimum size=1cm] at (9,0) {};
  \node at (9,0.75) {9};
  \node at (9,-0.75) {$C_1$};

  \node[draw, pattern=north east lines, pattern color=black, minimum size=1cm] at (10,0) {};
  \node at (10,0.75) {10};
  \node at (10,-0.75) {$C_2$};

  \node[draw, pattern=north west lines, pattern color=black, minimum size=1cm] at (11,0) {};
  \node at (11,0.75) {11};
  \node at (11,-0.75) {$\Sigma_C$};

  \node at (12.2, 0) {\ldots};
\end{tikzpicture}}
		\caption{The order of individuals in a population. The example presents the head of~a~population for three criteria: $C_0$, $C_1$, and $C_2$. Individuals are sorted according to fitness for each criterion and their sum, $\Sigma_C$. Four leading individuals of the population have the highest fitness according to individual criteria and their sum.}
		\label{fig:order}
	\end{figure}

\paragraph{Crossover}
The crossover is described in the lines \verb|10:17| of the pseudocode. The number of operations is calculated based on the crossover ratio $r_c$ and the population size $m$. At least one crossover will be performed in each iteration. The procedure combines one individual from the head of the population with another, randomly chosen. The result of a crossover is a weighted sum of transformation matrices with random weights. By default, a higher weight is associated with the better-adapted individual. The new individual replaces the worst-adapted one from the tail of the population.

\paragraph{Mutation}
The mutation is described with the lines \verb|18:23| of the pseudocode. The operation affects a randomly selected individual. Mutation involves adding random noise to each coefficient. The random values affecting the transformation are drawn from a~normal distribution $\mathcal{N}(0,0.1)$. The $\sigma=0.1$, which is significantly lower than the standard deviation of the distribution from which the individual's values are originally drawn ($\sigma=3$), means that the mutation only slightly affects the transformation matrix. After the crossover and mutation, the ratios $r_c$ and $r_m$ are reduced according to the decay factor $d_f$, which improves the stability of the search in the later evolution phase. With the significant decay factor and the number of iterations, the algorithm's operation in the later phases of processing will involve a single crossover per iteration.

\begin{figure}[!htb]
    \centering
    \includegraphics[width=\textwidth,trim={1.5cm 0 1.5cm 0},clip]{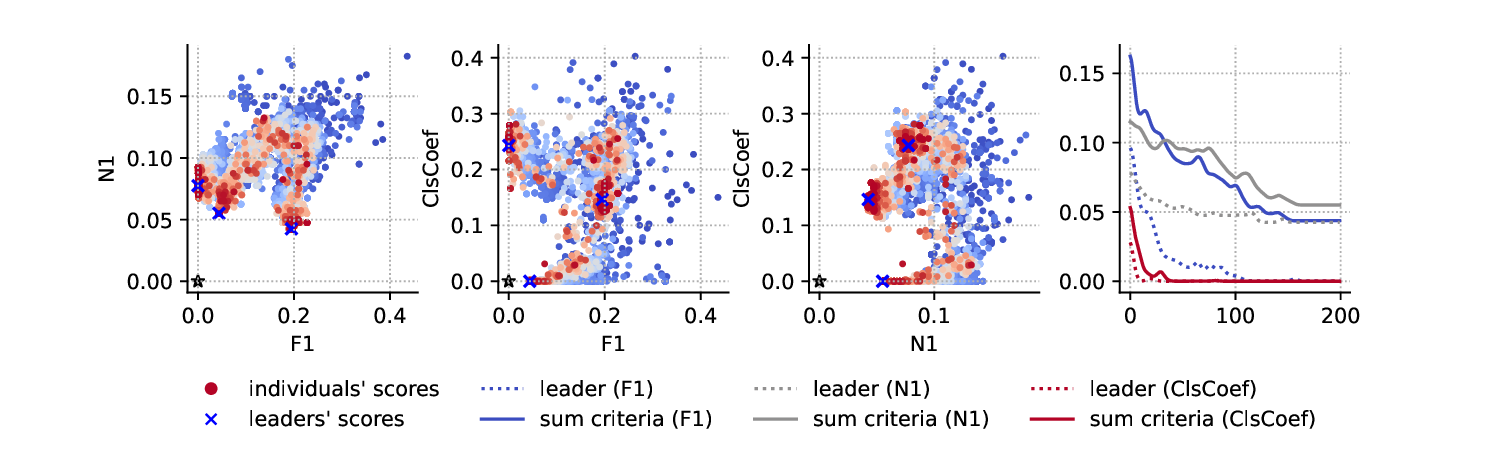}
    \caption{The optimization of three complexity measures -- F1, N1, and ClsCoef. The first 3 subfigures show the relationships between the criterion pairs. The individual points indicate the population's fitness at a given optimization stage. $X$ markers indicate leaders of the population. The last plot shows the leaders' scores across 200 iterations.}
    \label{fig:pareto}
\end{figure}

Figure \ref{fig:pareto} presents the optimization process of the \textsc{epco} algorithm for the exemplary set of three classification measures. The blue points indicate the scores of individuals at the initial stage of the optimization, while the red points show the final population. The star marker shows the target of optimization, indicating the achievement of a target value $T_i$ of a given complexity measure $C_i$. The last plot in the figure shows how the fitness of the population leaders evolved over the evolutionary process for each criterion.

The optimization loop continues until the iteration limit is reached. The result of an algorithm is a dataset $\{X_p, y\}$ in which the set of features $X_p$ results from the projection of original features' with a specific individual from the population. By default, the individual with the highest score in all criteria ($\Sigma_C$) is selected for projection. If requested, the algorithm can return a dataset transformed by any projection from the population. For example, using the projection $p_0$ in the transformation in line \verb|27| of the pseudocode will result in the selection of the individual with the highest fitness in the $C_0$ criterion. 

The proposal of a custom approach to optimization, rather than relying on NSGA-III \cite{deb2013evolutionary,fracca2020many}, was motivated by the potential for strong diversification within a relatively small population. In NSGA-III, the population is formed based on the \textit{reference points} spanning the \textit{Pareto frontier}. In contrast, \textsc{epco} optimized the individual criteria and their sum, serving as an equivalent of $|\mathcal{C}|+1$ reference points. In future research, the dataset generation approach relying on feature projections (characteristic of \textsc{epco}) can be extended to a more computationally costly but more precise algorithm. This research aims to present a simple yet effective solution to complexity optimization that relies on feature transformation, which is a novelty compared to methods known from the literature.

\begin{table}[t]
    \centering
    \caption{The description of the complexity measures selected for optimization with the \textsc{epco} algorithm}
    \setlength{\tabcolsep}{3pt}
    \scriptsize
    \begin{tabularx}{\textwidth}{c|l|X}
    \toprule
    \textsc{task} & \textsc{acronym} & \textsc{measure description}
    \\ \midrule
     \multirow{16}{*}{\rotatebox[origin=c]{90}{\textit{classification}}} & F1 & \textit{Maximum Fisher’s discriminant ratio} -- measures the class overlap \\
     & F3 & \textit{Maximum individual feature efficiency} -- measures the efficiency of each feature in the separation of classes\\
     & F4 & \textit{Collective feature efficiency} -- measures the features synergy in class separation \\
     & L2 & \textit{Error rate of linear classifier} -- calculated using linear SVM classifier \\
     & N1 & \textit{Fraction of borderline points} -- computed based on the number of edges in the \textit{minimum spanning tree} (\textsc{mst}) between examples of different classes \\
     & N3 & \textit{Error rate of NN classifier} -- calculated using the 1-nearest neighbor classifier \\
     & N4 & \textit{Non-linearity of NN classifier} -- the error rate of the k-Nearest Neighbor Classifier measured on synthetic samples \\
     & T1 & \textit{Fraction of hyperspheres covering data} -- measured by covering data samples with growing hyperspheres\\
     & ClsCoef & \textit{Clustering Coefficient}  -- calculated based on the number of edges between the sample’s neighbors in the graph generated based on the data \\
     & T4 & \textit{Ratio of the PCA dimension to the original dimension} -- computed based on the number of PCA components needed to represent 95\% of the data variance \\ \midrule
     \multirow{8}{*}{\rotatebox[origin=c]{90}{\textit{regression}}} & C1 & \textit{Maximum feature correlation to the output} -- maximal absolute value out of all feature-output correlations \\
     & C2 & \textit{Average feature correlation to the output} -- average absolute value out of all feature-output correlations \\
     & S1 & \textit{Output distribution} -- smoothness measure that calculates a difference in output values in a \textsc{mst} built on the data samples \\
     & S2 & Input distribution -- measures a difference in input values in a \textsc{mst} build on the data samples\\
    \end{tabularx}
    \label{tab:measures}
\end{table}

\subsection{Selected complexity measures}

The method was designed to transform the datasets for both classification and regression. Each of those tasks has a predefined set of complexity measures estimating the difficulty of a given dataset. The initially considered sets of complexity measures are consistent with those described by Lorena et al.~\cite{lorena2019complex,lorena2018data}, resulting in the analyses of 22 metafeatures for classification tasks and 12 for regression tasks. Out of the 22 designed for classification, two measures expressing class imbalance (C1 and C2) were dropped since the presented \textsc{epco} does not allow modification of the class proportions. The complete set of measures for each task was then examined to investigate the possibility of modifying them via feature projections. This resulted in the selection of ten measures for the classification task and four measures for the regression task, summarized in Table~\ref{tab:measures}.

The considered measures were selected in the preliminary single-criteria experiment, where the \textsc{epco} algorithm aimed to achieve different values of complexity from a range $[0:1]$. The experiment altered a single synthetic dataset per task. The measures for which a meaningful range of complexity values could be obtained were selected for further many-criteria experiments. The only measures not included, despite this observation, were C3 and C4 from the regression category. These two measures showed a strong correlation with C1 and C2 and were significantly more costly to compute.

\begin{table}[!b]
    \centering
    \caption{Target complexity measures for five various levels of complexity for classification and regression tasks -- from \textit{easy} to \textit{complex} target problems. The values are rounded to two decimal points.}
    \setlength{\tabcolsep}{3pt}
    \scriptsize
    \begin{tabularx}{\textwidth}{l|XXXXXXXXXX|XXXX}
    \toprule
    & \multicolumn{10}{c}{\textsc{classification}} & \multicolumn{4}{c}{\textsc{regression}}\\
    & F1 & F3 & F4 & L2 & N1 & N3 & N4 & T1 & ClsC & T4  & C1 & C2 & S1 & S2 \\ \midrule
    \textsc{easy}        & 0.20   & 0.45   & 0.00      & 0.05 & 0.05   & 0.10 & 0.05   & 0.60 & 0.45   & 0.50 & 0.90 & 0.40 & 0.10    & 0.90                \\
    \textsc{m/easy}    & 0.38 & 0.59 & 0.21 & 0.10  & 0.11 & 0.20 & 0.11 & 0.70 & 0.59 & 0.54 & 0.70 & 0.30 & 0.14 & 0.93            \\
    \textsc{medium}      & 0.55  & 0.73  & 0.43  & 0.15 & 0.18  & 0.30 & 0.18  & 0.80 & 0.73  & 0.58 & 0.50 & 0.20 & 0.18  & 0.95                   \\
    \textsc{m/complex} & 0.73 & 0.87 & 0.64 & 0.20  & 0.24 & 0.40 & 0.24 & 0.90 & 0.86 & 0.61 & 0.30 & 0.10 & 0.21 & 0.98                 \\
    \textsc{complex}     & 0.90   & 1.00      & 0.85   & 0.25 & 0.30    & 0.50 & 0.30    & 1.00   & 1.00      & 0.65 & 0.10 & 0.00   & 0.25   & 1.00  \\ 
    \bottomrule
    \end{tabularx}
    \label{tab:targets}
\end{table}

The target values describing five levels of problem complexity (from \textit{easy} to \textit{complex}) were also established based on the preliminary experiment and are presented in Table \ref{tab:targets}. All selected measures, except C1 and C2 for regression tasks, return higher values for more complex tasks. Hence, the target values for most measures increase when the goal is to make the problem more difficult, but for C1 and C2, the target values decrease. The measure values presented in the table were evenly sampled from a range established in the preliminary experiment. The presented measure values will serve as targets $\mathcal{T}$ in the experiments.

It is worth noting that the possibility of obtaining specific measure values depends on the source distribution. For example, in a synthetic regression dataset where the label is a linear combination of features, the measures of linearity (L1 and L2) will remain constant regardless of the data transformation. The measures selected in the preliminary experiment and their values, presented in Table \ref{tab:targets}, should serve as a starting point for the \textsc{epco} configuration, which can be altered with method hyperparameters to suit a particular type of data.

\section{Experiment design}

The presented algorithm was implemented in \textit{Python} and evaluated in a series of computer experiments. The code is publicly available\footnote{\url{https://github.com/JKomorniczak/EPCO}}. The complexity measures used in the research were calculated using the \textit{problexity} library \cite{komorniczak2023problexity}. The implementation of synthetic data generation methods and recognition methods came from \textit{scikit-learn} and used the default hyperparameters. All experiments used 5-times repeated 2-fold cross-validation.

\subsection{Source datasets}

For the optimization process, the synthetically generated datasets were used as a source distribution $\{X, y\}$. For that purpose, we used the \textit{Madelon} data generator (\verb|make_classification|) and the random regression data generator (\verb|make_regression|). For both tasks, the data were described by $20$ features, of which $2$ were informative for the binary classification and $10$ for the regression. For classification, the default label noise of $1\%$ was present, whereas for the regression, Gaussian noise with a standard deviation of $1.0$ was introduced. Both types of datasets consisted of $350$ samples. To stabilize the results, each of the presented experiments was performed in $10$ replications, with source datasets generated based on various random seeds.

\subsection{Algorithm configuration}

The configuration of the \textit{epco} algorithm was common for both evaluated tasks. The optimization was performed over $100$ iterations with $100$ individuals in the population. The crossover ratio was $25\%$, the mutation ratio $10\%$, and both were reduced throughout optimization with a decay factor of $0.7\%$. As mentioned, the number of criteria varied depending on the task, resulting in the optimization of $10$ measures for classification and $4$ for regression.

As a result of the optimization, $|\mathcal{C}| + 1$ datasets were constructed and saved, where $|\mathcal{C}|$ denotes the number of considered criteria. This resulted in $11$ classification datasets and $5$ regression datasets constructed based on each source distribution and for each target complexity level. The generated datasets were a result of transformation using population leaders. Such an operation will result in generating datasets best fitted to each individual of the considered criteria $C_i$ and to the sum of all criteria $\Sigma_C$. 

The generated datasets were saved for further evaluation, which resulted in $550$ classification datasets ($110$ for each of the $5$ target complexities) and $250$ regression datasets ($50$ per target complexity).

\subsection{Recognition methods and quality metrics}

After generating datasets with various complexities according to the selected measures, the final experiment aimed to directly assess the classification and regression qualities of the generated data using state-of-the-art methods. The evaluated classifiers included k-Nearest Neighbors (\textsc{knn}), Decision Tree (\textsc{dt}), Gaussian Naive Bayes (\textsc{gnb}), Multilayer Perceptron (\textsc{mlp}), and Support Vector Machine (\textsc{svm}). Since the generated datasets were balanced, the classification experiment used accuracy as the evaluation metric.

The evaluated regression methods included k-Nearest Neighbors Regressor (\textsc{knr}), Decision Tree Regressor (\textsc{dtr}), Bayesian Ridge Regressor (\textsc{brr}), Multilayer Perceptron Regressor (\textsc{mlpr}), and Support Vector Regressor (\textsc{svr}). In the regression experiments, the Mean Absolute Error (\textsc{mae}) was used as a quality metric.

\section{Experiment results}

This section presents the results of experiments evaluating both the \textit{epco} optimization capabilities and the recognition performance on classification and regression datasets. The datasets were generated using feature projections derived from the optimization of complexity measures towards five target complexities.

\begin{table}[!b]
    \centering
    \scriptsize
    \caption{Average complexity measure values for classification datasets. The values describe the complexity of a dataset generated using \textsc{epco} and -- in parentheses -- the average fitness function of the result (describing the absolute difference from the target) across all criteria.}

    \setlength{\tabcolsep}{6pt}
    \begin{tabularx}{\textwidth}{l|X|X|X|X|X}
    \toprule
    \textit{measure} & \textsc{easy}          & \textsc{m/easy}      & \textsc{medium}        & \textsc{m/complex}   & \textsc{complex}       \\ \midrule
 F1      & 0.307 (0.107) & 0.357 (0.020) & 0.512 (0.047) & 0.730 (0.007) & 0.885 (0.019) \\
 F3      & 0.606 (0.156) & 0.639 (0.052) & 0.773 (0.048) & 0.919 (0.057) & 0.967 (0.033) \\
 F4      & 0.097 (0.097) & 0.225 (0.024) & 0.424 (0.013) & 0.625 (0.013) & 0.747 (0.103) \\
 L2      & 0.071 (0.032) & 0.069 (0.034) & 0.083 (0.067) & 0.131 (0.069) & 0.166 (0.091) \\
 N1      & 0.087 (0.037) & 0.107 (0.007) & 0.149 (0.026) & 0.203 (0.035) & 0.227 (0.073) \\
 N3      & 0.164 (0.065) & 0.200 (0.011) & 0.289 (0.013) & 0.401 (0.009) & 0.457 (0.050) \\
 N4      & 0.051 (0.007) & 0.068 (0.016) & 0.111 (0.023) & 0.181 (0.013) & 0.230 (0.044) \\
 T1      & 0.815 (0.215) & 0.871 (0.171) & 0.938 (0.138) & 0.981 (0.081) & 0.993 (0.007) \\
 ClsCoef & 0.609 (0.159) & 0.629 (0.043) & 0.729 (0.021) & 0.841 (0.024) & 0.845 (0.155) \\
 T4      & 0.530 (0.030) & 0.555 (0.018) & 0.565 (0.025) & 0.600 (0.013) & 0.595 (0.055) \\
    \bottomrule
    \end{tabularx}
    \label{tab:classification}
\end{table}

\subsection{Classification}

Table \ref{tab:classification} presents the complexities of the output datasets after optimization using \textsc{epco}. The results show average complexities for the projection with the best fitness in the sum of the criteria ($\Sigma_C$) -- which is the default result of the algorithm. The value in parentheses can be interpreted as an error -- the distance from the target to the obtained value.

The results show that most of the target metric values were possible to obtain. The largest errors are seen for the T1 measure across all target difficulties, and for the F1, F3, and ClsCoef measures at the extremes of the possible target ranges (\textit{easy} and \textit{complex} tasks). The inability to obtain measures at the lower or upper ends of the selected range may result from the optimization considering multiple criteria. The values that were achievable for a single metric in the preliminary experiment -- based on which the ranges were selected -- may become unreachable when another measure is considered equally important. Even though some differences from the targets are observed, the direct values of complexities increase with the expected difficulty for most measures.

\begin{figure}[!b]
    \centering
    \includegraphics[width=\textwidth]{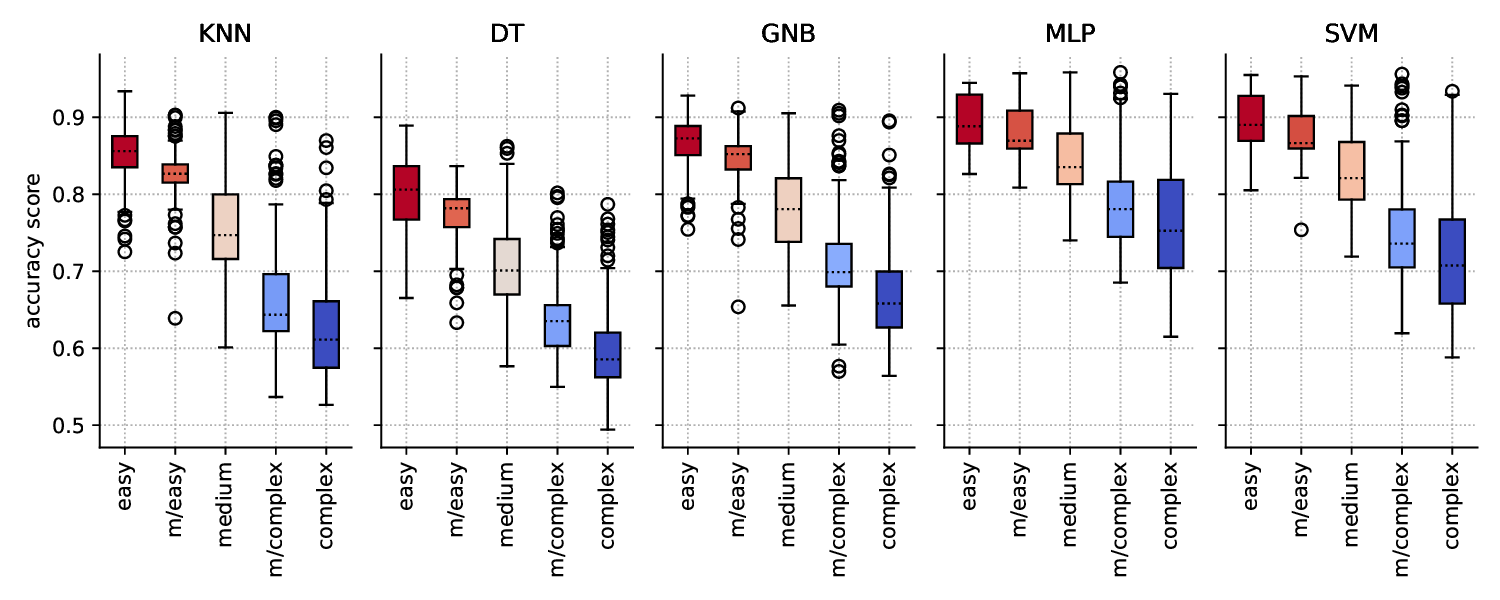}
    \caption{The results of the classification accuracy for datasets transformed with \textsc{epco} towards various difficulty levels. The larger complexity of the data resulted in lower accuracy scores for all evaluated classifiers.}
    \label{fig:clf}
\end{figure}

Another experiment for classification used a larger number of datasets returned by the \textsc{epco} method (a total of $550$ constructed based on $11$ population leaders) and evaluated the recognition quality using five classification methods. The results of classification accuracy obtained in this experiment are presented in Figure \ref{fig:clf}. The columns show the achieved scores for specific classifiers, with box colors indicating their average quality. Reds correspond to the higher accuracy, and blues correspond to the lower scores. The bars represent the results for various requested difficulties -- from \textit{easy} to \textit{complex}.

The results of this experiment indicate that the datasets with higher complexity are more difficult in terms of the recognition task. This is observed for all the evaluated classifiers. Interestingly, classification quality seems to degrade linearly as complexity increases. Since the target values of the complexity measures used in the \textsc{epco} algorithm were also sampled using a linear function in the selected ranges, one may expect that the classification quality is proportional to the value of the complexity measure.

\subsection{Regression}

Table \ref{tab:regression} presents the average dataset complexity generated by the \textsc{epco} algorithm for the regression task. For regression problems, only 4 measures were used. Since the C1 and C2 measures for regression tasks return lower values for more complex problems, their values decrease when the expected difficulty of the dataset is intended to increase. Similarly to the classification results table, measures were calculated for the individual that best fit the sum of the criteria ($\Sigma_C$).

\begin{table}[!t]
    \centering
    \scriptsize
    \caption{Average optimization results for regression task. The values describe the complexity obtained as a result of \textsc{epco} search and -- in parentheses -- the fitness function across all criteria.}

    \setlength{\tabcolsep}{6pt}
    \begin{tabularx}{\textwidth}{l|X|X|X|X|X}
    \toprule
    \textit{measure} & \textsc{easy}          & \textsc{m/easy}      & \textsc{medium}        & \textsc{m/complex}   & \textsc{complex}       \\ \midrule
      C1 & 0.884 (0.017) & 0.699 (0.001) & 0.500 (0.002) & 0.300 (0.001) & 0.135 (0.035) \\
     C2 & 0.278 (0.122) & 0.295 (0.006) & 0.202 (0.002) & 0.102 (0.003) & 0.075 (0.075) \\
     S1 & 0.102 (0.007) & 0.106 (0.031) & 0.147 (0.028) & 0.186 (0.027) & 0.192 (0.058) \\
     S2 & 0.969 (0.069) & 0.957 (0.032) & 0.982 (0.032) & 1.027 (0.052) & 1.044 (0.044) \\
    \bottomrule
    \end{tabularx}
    \label{tab:regression}
\end{table}

In the case of regression tasks, some measures show a stable decrease (C1 and C2) or increase (S1 and S2) when raising the target difficulty. The C1 measure was the easiest to optimize -- its fitness in parentheses describes the smallest differences from the target. C2 for the \textit{easy} target is the only case in which the optimization error exceeded 0.1. However, as mentioned, targets lying at the extremes of the possible range may be more challenging to obtain. 

\begin{figure}[!b]
    \centering
    \includegraphics[width=\textwidth]{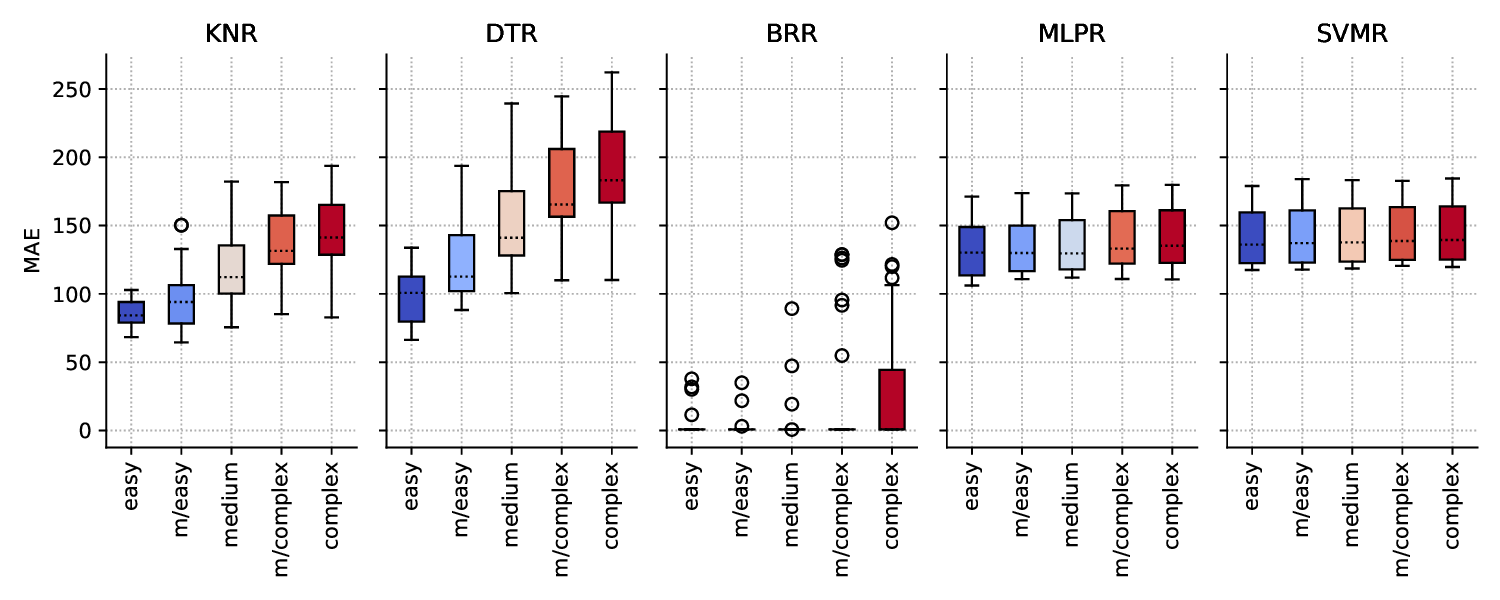}
    \caption{The results of the \textit{mean absolute error} obtained by the evaluated regression methods for datasets transformed with \textsc{epco} towards target difficulties. More challenging problems are associated with higher recognition errors.}
    \label{fig:reg}
\end{figure}

Similarly to the classification experiment, the regression data sets were also subjected to additional analysis that examined the recognition quality. This experiment was carried out on a total of $250$ datasets generated using the \textsc{epco} algorithm. Results are presented in Figure \ref{fig:reg}. The figure shows how evaluated regression methods handle datasets with various complexities by presenting the values of \textsc{mae}.

In the regression task, one can observe an increase in errors for the evaluated methods as the data difficulty increases. The errors differ significantly depending on the target complexity in the case of the \textsc{knr} and \textsc{dtr} methods. In the case of \textsc{mlpr} and \textsc{svmr}, the differences in quality are also visible -- highlighted by the changes in the colors of the boxes -- however, the range of obtained \textsc{mae} is smaller compared to the remaining methods. In the case of \textsc{brr}, one can mostly notice the increasing errors of outliers indicated by individual points. The low errors of this method may result from the generation of the original data using the \verb|make_regression| function, which models a linear relationship between features and a target.

\section{Conclusions}

This work aims to increase the availability of synthetic datasets characterized by a wide range of problem complexities. Fulfilling this goal may supplement the reliable and fair evaluation of machine learning methods.

To alter the complexity of synthetically generated datasets, a set of 10 complexity measures for classification and 4 for regression tasks has been optimized in the proposed \textsc{epco} approach. The proposed generic algorithm searches the space of feature projection matrices, which are used to transform the source data distribution towards a target complexity profile. The method was used to generate $550$ classification datasets and $250$ regression datasets towards five different target difficulty levels. Experiments evaluating state-of-the-art classifiers and regression methods showed a decrease in classification accuracy and an increase in regression errors when faced with datasets of higher complexity.

The presented algorithm used a custom genetic approach to many-criterion optimization, where individuals in the population were sorted according to fitness to the considered criteria and their sum. This simple approach yielded promising optimization results, especially given a small population and up to ten considered criteria. Future work may employ more sophisticated optimization strategies, in which the final individuals of the population form a more continuous \textit{Pareto frontier} of the optimized complexity measures. Studying more complex transformation strategies is also a promising future research direction.

\begin{credits}
\subsubsection{\ackname} 
This work was supported by the statutory funds of the Department of Systems and Computer Networks, Faculty of Information and Communication Technology, Wrocław University of Science and Technology.

\subsubsection{\discintname}
The authors have no competing interests to declare that are
relevant to the content of this article.
\end{credits}

\bibliographystyle{splncs04}
\bibliography{bib.bib}

\end{document}